%% file: main.tex
\documentclass[journal,twoside,web]{ieeecolor}
\usepackage{generic}
\usepackage{cite}
\usepackage{amsmath,amssymb,amsfonts}
\usepackage{algorithmic}
\usepackage{graphicx}
\usepackage{algorithm,algorithmic}
\usepackage{hyperref}
\hypersetup{hidelinks=true}
\usepackage{textcomp}
\usepackage{booktabs}
\usepackage{placeins}
\usepackage{xcolor}
\usepackage{colortbl}
\usepackage{multirow}
\definecolor{TableBest}{HTML}{A52F2B}
\definecolor{TableSecond}{HTML}{557D96}
\newcommand{\rankfirst}[1]{\textcolor{TableBest}{\mathbf{#1}}}
\newcommand{\ranksecond}[1]{\textcolor{TableSecond}{\underline{#1}}}
\usepackage{hyperref}
\hypersetup{hidelinks}
\usepackage{etoolbox}
\makeatletter
\def\BibTeX{{\rm B\kern-.05em{\sc i\kern-.025em b}\kern-.08em
    T\kern-.1667em\lower.7ex\hbox{E}\kern-.125emX}}
\begin{document}
\title{Mitral-AIM: Interval-Aware Anatomy--Motion Learning for Mitral Regurgitation Detection from B-Mode~4D~TEE}
\author{Wenlong Hou, Tsz Wai Wong, Mingjun Yin, Juncheng Wang, Guoqi Yu, Nanxi Yu, Qian Niu, Jing Qin, \IEEEmembership{Member, IEEE}, Shing Ching, Shujun Wang, \IEEEmembership{Member, IEEE}
\thanks{*Corresponding authors: Shing Ching and Shujun Wang.}
\thanks{Wenlong Hou, Tsz Wai Wong, Mingjun Yin, Juncheng Wang, Guoqi Yu, Nanxi Yu, Jing Qin, and Shujun Wang are with the Hong Kong Polytechnic University, Hong Kong SAR, China
(e-mails: \{willen-wenlong.hou, tszwai15.wong, 24104771d\}@connect.polyu.hk, wjc2830@gmail.com, \{levi-ack.yu, nx-nancy.yu\}@connect.polyu.hk, \{harry.qin, shu-jun.wang\}@polyu.edu.hk).}
\thanks{Qian Niu is with the University of Tokyo, Japan (e-mail: qian.niu@weblab.t.u-tokyo.ac.jp).
Shing Ching is with the United Christian Hospital, Hong Kong SAR, China (e-mail: aj12029@yahoo.com.hk).}}

\maketitle
\thispagestyle{plain}
\input{sections/abstract}

\input{sections/keywords}

\input{sections/introduction}
\input{sections/related_work}
\input{sections/materials_and_methods}
\input{sections/results}
\input{sections/discussion}
\input{sections/conclusion}
\FloatBarrier

\bibliographystyle{IEEEtran}
\bibliography{references}
\end{document}

%% file: sections/abstract.tex
\begin{abstract}
Mitral regurgitation (MR) results from incomplete mitral valve closure during systole and is associated with heart failure and increased mortality. MR assessment typically relies on Doppler measurements of regurgitant flow. However, accurate acquisition and interpretation of these measurements require substantial clinical expertise. Therefore, we present Mitral-AIM, to our knowledge the first deep learning framework to detect clinically significant MR directly from B-mode four-dimensional transesophageal echocardiography (4D TEE) sequences without Doppler input. The mitral valve occupies a small portion of each 4D TEE frame, temporal changes between consecutive frames are subtle, and systole is particularly informative for MR detection. 
To address these challenges, Mitral-AIM combines (i) \textit{local--global anatomical encoding} to capture local detail and global context; (ii) \textit{inter-volume motion encoding} to independently extract motion cues from frame-to-frame intensity changes; and (iii) \textit{interval-based temporal weighting} to learn diagnostically informative intervals. Mitral-AIM is trained using only patient-level diagnostic labels, without valve segmentation or cardiac-phase annotations.
Experiments on a privately collected dataset showed that Mitral-AIM achieved the highest patient-level AUROC among the evaluated methods ($76.7\pm1.9\%$), 4.4 percentage points above the best baseline. Further analyses demonstrated that the learned temporal weighting aligned with systole and highlighted intervals important for diagnostic performance. Representation analysis further revealed that both local and global anatomical features captured variation across cardiac phases. The source code will be available at \url{https://github.com/SDH-Lab/Mitral-AIM}.
\end{abstract}

%% file: sections/keywords.tex

%% file: sections/introduction.tex
\section{Introduction}\label{sec:introduction}
\IEEEPARstart{M}{itral} regurgitation (MR) occurs when ineffective systolic coaptation allows blood to flow backward from the left ventricle into the left atrium (Fig.~\ref{fig:mitraiam_teaser}(A)) \cite{zoghbi2017}. MR may result from intrinsic abnormalities of the leaflets or subvalvular apparatus, or from left ventricular or left atrial remodeling that disrupts leaflet coaptation \cite{zoghbi2017,robinson2021}. The global burden is substantial: non-rheumatic degenerative mitral valve disease alone affected an estimated 15.5 million people and accounted for approximately 36,800 deaths and 943,000 disability-adjusted life years in 2021 \cite{chen2024global,kong2025global}. Accurate and timely identification is therefore important for surveillance, etiological assessment, and treatment planning.

\input{figures/mitraiam_teaser}

Echocardiography is the principal imaging modality for assessing MR.
As illustrated in Fig.~\ref{fig:mitraiam_teaser}(B), Doppler-based assessment entails three key requirements:
(1) integration of color and spectral Doppler measurements to characterize regurgitant flow;
(2) careful optimization of acquisition parameters, including color gain and velocity scale; and
(3) clinical judgment in the interpretation of potentially discordant findings \cite{zoghbi2017,robinson2021}.
These requirements motivate the investigation of B-mode imaging as an alternative source of diagnostic information. Although B-mode imaging does not directly quantify regurgitant flow, it depicts valve anatomy and its temporal evolution. Accordingly, we investigate automated detection of clinically significant MR from B-mode four-dimensional transesophageal echocardiography (4D TEE) sequences, using grayscale volumes without Doppler data as model input.


B-mode 4D TEE captures a time-resolved sequence of three-dimensional grayscale cardiac volumes, enabling mitral valve morphology and its changes over the cardiac cycle to be examined. Its volumetric representation provides access to leaflet configuration, annular geometry, and their relationships with surrounding structures \cite{faletra2023}. Clinically significant MR is associated with distinct morphologic substrates that can be characterised using B-mode 4D TEE. For example, leaflet perforation, leaflet thickening, or chordal rupture can be observed in primary MR \cite{zoghbi2017}; secondary MR can involve leaflet tenting, annular dilation, and impaired coaptation \cite{robinson2021}. These structural abnormalities and their temporal expression provide a clinical rationale for investigating MR-associated anatomical and dynamic information in B-mode volume sequences. We investigate whether joint analysis of valve morphology, surrounding anatomy, and their temporal evolution can distinguish clinically significant MR.

All existing learning-based MR assessment relies solely on two-dimensional transthoracic echocardiography (TTE), using either B-mode or color Doppler cine loops. No MR assessment model exists for TEE, let alone for B-mode 4D TEE. DELINEATE-MR selects color Doppler videos from TTE examinations and aggregates video-level severity predictions into a study-level grade \cite{long2024}. MitralVision detects clinically significant MR from single-view apical four-chamber B-mode cine loops without Doppler input \cite{sandler2026}. For volumetric echocardiography, including 3D and 4D TEE, learning-based studies have primarily focused on mitral valve segmentation and geometric analysis \cite{munafo2024,munafo2025,chen2025mtcnet}. Other work combines image quality assessment, valve segmentation, and anatomical quantification to support intervention planning \cite{corona2025}. These volumetric studies characterize valve anatomy but do not directly learn MR status from B-mode volume sequences. Direct detection of clinically significant MR from B-mode 4D TEE sequences thus remains underexplored.

As outlined in Fig.~\ref{fig:mitraiam_teaser}(C), learning this diagnostic mapping from patient-level labels poses three challenges.
First, fine-scale details of mitral valve coaptation are spatially localized within a large volume, making these local valve features susceptible to loss during spatial downsampling and global pooling.
Capturing these features requires preserving fine anatomical detail alongside the surrounding structural context. Second, changes between adjacent volumes can be subtle, requiring representations that capture inter-volume variation beyond the anatomical appearance of individual frames. Third, systole is particularly informative for assessing leaflet coaptation, whereas patient-level labels provide no indication of when relevant abnormalities occur. The model must therefore identify and aggregate informative temporal intervals within the full cardiac cycle without cardiac-phase annotations. Attention pooling~\cite{ilse2018} and TDA-Transformer~\cite{ahmadi2024} can learn frame importance under diagnostic supervision, but do not necessarily organize that evidence into coherent temporal intervals. This motivates adaptive temporal weighting that incorporates contiguous-interval structure within variable-length sequences, without cardiac-phase or frame-level diagnostic annotations. These requirements call for a representation preserving local anatomy, explicit inter-volume change, and coordinated aggregation weights for both streams.

To address these challenges, we propose Mitral-AIM, a weakly supervised framework for interval-aware anatomy--motion learning.
Its local--global anatomical encoding module integrates spatially attended local features with global features, while a complementary inter-volume motion encoding module independently encodes intensity differences between adjacent volumes.
An interval-based temporal weighting module then jointly uses anatomical and motion information to learn the diagnostic importance of each frame while accounting for temporal continuity.
The learned weights guide the aggregation of both anatomical and motion features.
These aggregated representations are combined with sequence-wide anatomical context to predict MR status.


The main contributions of this study are as follows:
\begin{enumerate}
\item To the best of our knowledge, we present the first study of detecting clinically significant MR directly from B-mode 4D TEE without Doppler input. We curate a private dataset of 623 volumetric sequences and benchmark 30 baseline configurations spanning conventional machine learning, volumetric and video models, multiple-instance learning, and echo-pretrained encoders.
\item We propose Mitral-AIM, which combines local--global anatomical features with explicit inter-volume motion encoding. Its interval-based temporal weighting module learns a distribution over contiguous temporal intervals to coordinate anatomical and motion aggregation.
\item We evaluate Mitral-AIM along diagnostic performance, temporal evidence, and anatomical representation. Comparisons with the benchmarked baselines show not only state-of-the-art (SOTA) performance but also temporal and anatomical evidence consistent with clinical logic.
\end{enumerate}

The remainder of this paper is organized as follows. Section~\ref{sec:related} reviews related work. Section~\ref{sec:methods} describes the dataset, proposed framework, and evaluation protocol. Section~\ref{sec:results} presents the experimental results and analyses. Section~\ref{sec:discussion} discusses the findings, and Section~\ref{sec:conclusion} concludes the paper.

%% file: figures/mitraiam_teaser.tex
\begin{figure}[!t]
\centering
\includegraphics[width=\columnwidth]{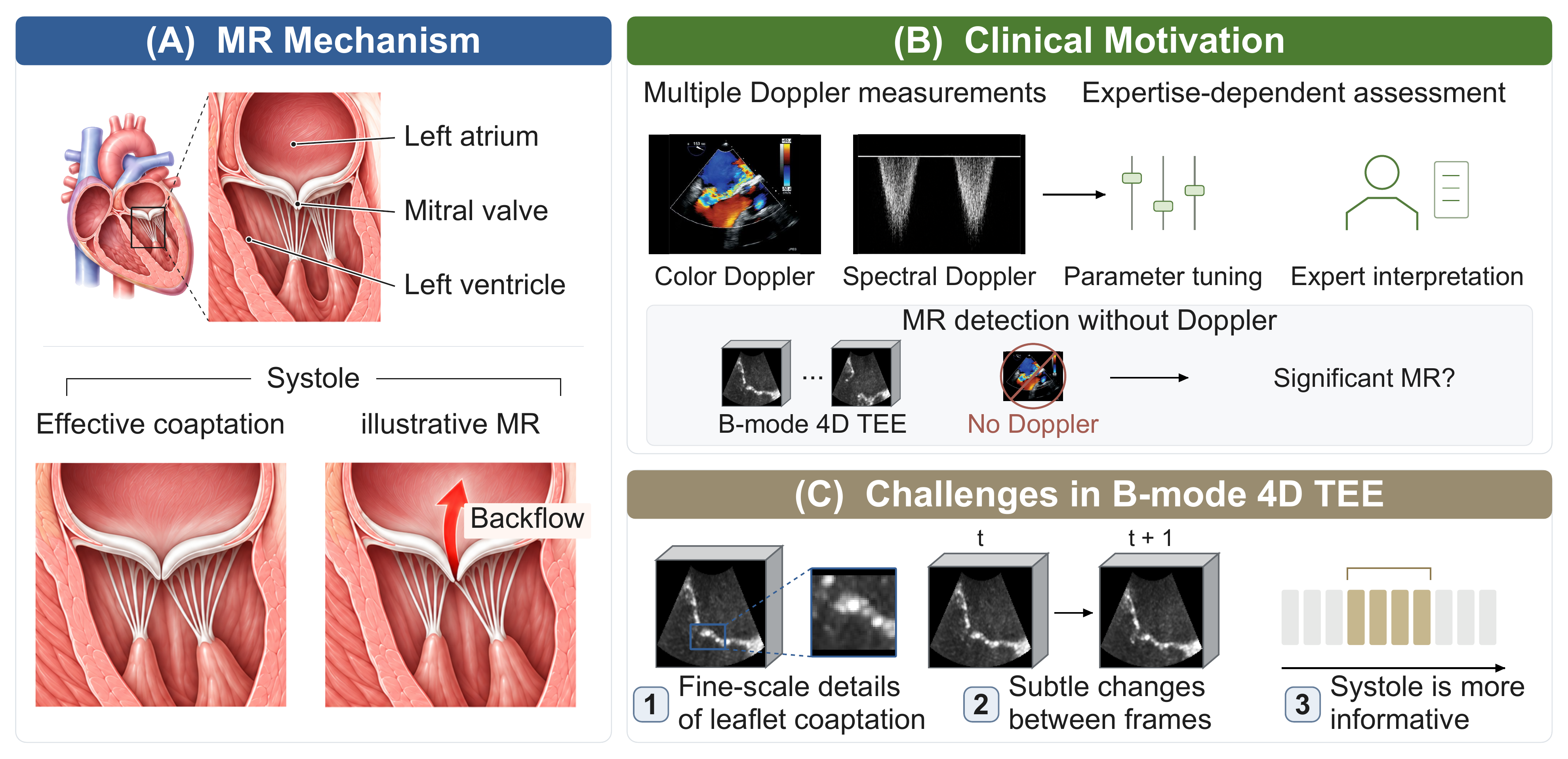}
\caption{Clinical motivation and challenges of MR detection from B-mode 4D TEE.
(A) Incomplete mitral valve closure during systole causes regurgitant flow into the left atrium.
(B) Doppler-based assessment requires multiple measurements and clinical expertise, motivating MR detection from B-mode 4D TEE alone~\cite{radiologykeyColorDoppler}.
(C) Key challenges include capturing fine-scale details of leaflet coaptation, distinguishing subtle inter-frame changes, and identifying diagnostically informative temporal intervals.}

\label{fig:mitraiam_teaser}
\end{figure}

%% file: sections/related_work.tex
\section{Related Work}
\label{sec:related}

\subsection{Learning-Based MR and Mitral Valve Analysis}

Learning-based MR assessment has been investigated using different echocardiographic inputs. In two-dimensional TTE, DELINEATE-MR identifies relevant color Doppler sequences, estimates sequence-level severity, and integrates predictions into a study-level grade~\cite{long2024}. MitralVision detects clinically significant MR from single-view apical four-chamber B-mode cine loops without Doppler input, establishing a precedent for direct MR detection from two-dimensional grayscale videos~\cite{sandler2026}.

In volumetric TEE, representative learning-based studies have focused on anatomical characterization. Munaf\`o et al. developed an automated pipeline for annular and leaflet segmentation, landmark detection, and geometric measurements from 3D TEE~\cite{munafo2024}. Their subsequent teacher--student framework extended segmentation across the cardiac cycle using pseudo-labels for intermediate volumes~\cite{munafo2025}. These approaches recover valve anatomy and its temporal evolution through segmentation-based analysis.

\subsection{Spatiotemporal Representation Learning}

Natural video recognition provides foundational approaches to modeling appearance and dynamics. Two-stream ConvNets separately encode images and optical flow~\cite{simonyan2014twostream}, while Temporal Segment Networks combine sparsely sampled snippets under video-level supervision and investigate input modalities including RGB differences~\cite{wang2016tsn}. These methods establish precedents for complementary appearance and motion-sensitive representations.

In echocardiography, EchoNet-Dynamic demonstrated video-based cardiac function assessment~\cite{ouyang2020echonet}, and Ultrasound Video Transformers introduced temporal attention for variable-length sequences, ejection fraction estimation, and cardiac keyframe detection~\cite{reynaud2021uvt}. TDA-Transformer addressed subtle valve dynamics using temporal deformable attention and a temporal coherence loss for aortic stenosis classification~\cite{ahmadi2024}. SwiFT applies shifted-window attention jointly over space and time to volumetric sequences~\cite{kim2023swift}.

Recent approaches emphasize transferable representations: EchoFM incorporates cardiac periodicity into masked video pretraining~\cite{kim2025echofm}, DISCOVR learns video representations through cluster distillation~\cite{mishra2025discovr}, EchoPrime combines video--report alignment with multiview analysis~\cite{vukadinovic2026echoprime}, and EchoJEPA uses latent predictive pretraining~\cite{munim2026echojepa}. This progression spans explicit motion encoding, task-specific temporal modeling, and large-scale echocardiographic pretraining.

\subsection{Weakly Supervised Temporal Evidence Aggregation}

Beyond representation learning, weakly supervised aggregation determines how individual observations contribute to prediction. Attention-based multiple instance learning learns instance weights from bag-level labels~\cite{ilse2018}. Later aggregators include DSMIL~\cite{li2021dsmil}, TransMIL~\cite{shao2021transmil}, CLAM-SB~\cite{lu2021clam}, DTFD-MIL~\cite{zhang2022dtfdmil}, ILRA~\cite{xiang2023ilra}, ACMIL~\cite{zhang2024acmil}, WiKG~\cite{li2024wikg}, and ASMIL~\cite{ye2026asmil}. In natural videos, Sparse Temporal Pooling Networks learn selective segment attention from video-level labels and generate temporal action proposals without boundary annotations~\cite{nguyen2018stpn}. These approaches connect coarse supervision with the identification of informative observations and temporal regions.

Cardiac studies address both physiological correspondence and diagnostic relevance. Echo-SyncNet learns self-supervised representations for temporal synchronization across echocardiographic views~\cite{dezaki2021echosyncnet}. TDA-Transformer learns diagnosis-driven frame weights, with an entropy penalty encouraging selective aggregation~\cite{ahmadi2024}. These objectives distinguish identifying corresponding cardiac states from weighting observations for disease prediction.


%% file: sections/materials_and_methods.tex
\section{Materials and Methods}
\label{sec:methods}
\input{figures/mitraiam_overview}
\subsection{Dataset and Study Population}
\label{sec:methods:dataset}
We investigate the detection of clinically significant MR from B-mode 4D TEE. The dataset comprises 623 volumetric sequences from 264 patients, acquired using a Vivid E95 ultrasound system (GE Vingmed Ultrasound). Each sequence represents a recorded cardiac cycle. Dataset characteristics are summarized in Table~\ref{tab:dataset_characteristics}. The study was approved by the ethics review board of the Hong Kong Hospital Authority (CIRB-2026-360-3), and was conducted according to the Declaration of Helsinki. The classification task is binary and distinguishes clinically significant MR from nonsignificant MR.


\input{tables/dataset_characteristics}

\subsection{Data Representation and Problem Formulation}
\label{sec:methods:formulation}

A sequence is represented as
$\mathcal{X}=\{X_t\}_{t=1}^{T},\qquad X_t\in\mathbb{R}^{D\times H\times W},$
where $X_t$ is a grayscale volume and $T$ denotes the number of acquired volumes. The learning objective is to estimate $p=f_{\theta}(\mathcal{X})$, where $p$ is the probability of clinically significant MR.

Each volume is resized to $64\times64\times64$ using trilinear interpolation. The same spatial mapping is applied to all volumes within a sequence, after verifying that their acquisition geometry is consistent. Intensity normalization uses a single scale computed from the 99th percentile of positive voxel intensities across the entire resized sequence, followed by clipping to $[0,1]$. Applying a shared normalization scale preserves inter-volume intensity relationships more directly than independently normalizing individual volumes.

All acquired volumes are retained in their original temporal order, without temporal interpolation, smoothing, or fixed-length frame sampling. During batching, shorter sequences are padded, and validity masks exclude padded positions from temporal attention and feature aggregation.

\subsection{Overview of Mitral-AIM}
\label{sec:methods:overview}
Fig.~\ref{fig:mitraiam_overview} presents the overall architecture of Mitral-AIM. Mitral-AIM learns complementary anatomical and motion-sensitive representations and integrates them using adaptive temporal aggregation weights derived from a shared interval distribution. Three core modules implement this integration. \textbf{\emph{Local--Global Anatomical Encoding}} combines spatially attended intermediate-resolution features with deep global features to describe each volume. \textbf{\emph{Inter-Volume Motion Encoding}} independently processes signed and absolute differences between adjacent volumes, providing an explicit representation of volumetric change. Joint Interval Aggregation contextualizes the two feature sequences. Its \textbf{\emph{Interval-Based Temporal Weighting}} module uses anatomy and motion to estimate a distribution over contiguous intervals and converts that distribution into coordinated weights, which are applied by weighted temporal pooling. The temporally weighted representations are combined with full-sequence anatomical context to produce the final prediction.

\subsection{Local--Global Anatomical Encoding}
\label{sec:methods:anatomy}
A shared 3D ResNet-18 encoder~\cite{hara2018} extracts complementary local and global anatomical features from each valid volume. Before encoding, the normalized input is transformed as $\widetilde{X}_t=(X_t-0.15)/0.25$. An intermediate feature map $F_t^{\mathrm{loc}}\in\mathbb{R}^{128\times8\times8\times8}$ is taken from the second residual stage and a deep feature map $F_t^{\mathrm{glob}}\in\mathbb{R}^{512\times2\times2\times2}$ from the fourth residual stage. Local attention scores are $e_{t,u}=g_A(F_{t,u}^{\mathrm{loc}})$, with GELU between the two $1\times1\times1$ convolutions in $g_A$.

To restrict attention to the ultrasound imaging region, we average-pool the input-support indicator $S_t(v)=\mathbf{1}[X_t(v)>0]$ to the resolution of $F_t^{\mathrm{loc}}$, obtaining support fractions $s_{t,u}\in[0,1]$. Attention weights are then computed as:
\begin{equation}
a_{t,u}=\frac{s_{t,u}\exp(e_{t,u})}{\sum_v s_{t,v}\exp(e_{t,v})}.
\label{eq:methods:spatial_attention}
\end{equation}
Locations with zero support receive zero attention weight.

The local and global anatomical descriptors are
$\ell_t=\phi_{\mathrm{loc}}\left(\sum_u a_{t,u}F_{t,u}^{\mathrm{loc}}\right)$, $g_t=\phi_{\mathrm{glob}}\left(\operatorname{GAP}(F_t^{\mathrm{glob}})\right)$,
where $\operatorname{GAP}$ denotes global average pooling, and $\phi_{\mathrm{loc}}$ and $\phi_{\mathrm{glob}}$ are linear projections followed by layer normalization, each producing a $d=256$ descriptor. The anatomical token is obtained by concatenating the two descriptors:
\begin{equation}
A_t=\operatorname{LN}\!\left[\operatorname{GELU}\!\left(W_A[\ell_t;g_t]+b_A\right)\right]\in\mathbb{R}^{d}.
\label{eq:methods:anatomy_token}
\end{equation}

\subsection{Inter-Volume Motion Encoding}
\label{sec:methods:motion}
Temporal modeling of anatomical tokens requires inter-volume changes to remain accessible after spatial feature extraction and pooling. To provide a complementary route for representing these changes, the motion stream operates directly on adjacent input volumes.

For each valid transition, we construct
\begin{equation}
\Delta X_t=X_{t+1}-X_t,\qquad t=1,\ldots,T-1,
\label{eq:methods:difference}
\end{equation}
and form a two-channel input $R_t=[\Delta X_t;|\Delta X_t|]$. The signed channel retains the polarity of intensity changes, while the absolute channel explicitly represents their magnitude.

A dedicated 3D convolutional encoder maps $R_t$ to a spatial motion representation. It contains three $3\times3\times3$ convolutional blocks with stride 2 and output channel counts of 32, 64, and 128. Each block uses group normalization and GELU activation. Two subsequent $1\times1\times1$ projection blocks, also followed by group normalization and GELU, produce $F_t^M\in\mathbb{R}^{256\times8\times8\times8}$. The motion encoder does not share parameters with the anatomical encoder. A separate spatial attention head aggregates $F_t^M$. Its support prior is the average support fraction of the two contributing volumes, $s_{t,u}^{M}=(s_{t,u}+s_{t+1,u})/2$. Using the same support-weighted normalization as in the anatomical stream, the motion attention weights and token are
\begin{align}
b_{t,u}&=\frac{s_{t,u}^{M}\exp(g_M(F_{t,u}^{M}))}{\sum_v s_{t,v}^{M}\exp(g_M(F_{t,v}^{M}))},\label{eq:methods:motion_attention}\\
M_t&=\sum_u b_{t,u}F_{t,u}^{M}\in\mathbb{R}^{d}.\label{eq:methods:motion_token}
\end{align}
Stacking tokens in acquisition order yields anatomical and motion feature sequences $A\in\mathbb{R}^{T\times d}$ and $M\in\mathbb{R}^{(T-1)\times d}$.

\subsection{Joint Interval Aggregation}
\label{sec:methods:interval}
\subsubsection{Temporal contextualization}
The anatomical and motion sequences are processed by independent temporal Transformers~\cite{vaswani2017}. To accommodate variable sequence lengths, we normalize anatomical token positions to $[0,1]$ and place each motion token at the midpoint of its contributing volumes: $\tau_t=\frac{t-1}{T-1}$, $\qquad \tau_t^{M}=\frac{\tau_t+\tau_{t+1}}{2}$. Both streams use the same fixed Fourier map to encode their positions on a shared temporal coordinate: $\gamma(\tau)=\left[\{\sin(2\pi f_k\tau)\}_{k=1}^{d/2};\{\cos(2\pi f_k\tau)\}_{k=1}^{d/2}\right]$, where $f_k=32^{(k-1)/(d/2-1)}$ for $k=1,\ldots,d/2$. The frequencies span 1--32 cycles over the normalized sequence, providing positional information at multiple temporal scales without trainable parameters. These encodings are added to the corresponding tokens before temporal contextualization:
\begin{align}
H^A&=\mathcal{T}_A\left(\{A_t+\gamma(\tau_t)\}_{t=1}^{T}\right),\label{eq:methods:anatomy_temporal}\\
H^M&=\mathcal{T}_M\left(\{M_t+\gamma(\tau_t^M)\}_{t=1}^{T-1}\right).\label{eq:methods:motion_temporal}
\end{align}
Each Transformer contains two pre-normalization layers, four attention heads, a model dimension of 256, and a feed-forward dimension of 512. Padded positions are masked.

\subsubsection{Interval-based temporal weighting}
This module learns aggregation weights for anatomical time points and inter-volume transitions from a shared distribution over contiguous intervals. Transitions are then mapped to volumes:
\begin{equation}
\widetilde{H}_t^M=\begin{cases}
H_1^M,&t=1,\\[1mm]
(H_{t-1}^M+H_t^M)/2,&1<t<T,\\[1mm]
H_{T-1}^M,&t=T.
\end{cases}
\label{eq:methods:alignment}
\end{equation}
The aligned features are concatenated and fused as $U_t=\phi_F([H_t^A;\widetilde{H}_t^M])$, where $\phi_F$ comprises a 512-to-256 linear projection, GELU, dropout, and layer normalization. Two independent MLP heads produce start and end scores, $s_t=g_s(U_t)$ and $e_t=g_e(U_t)$; each comprises layer normalization, a 128-unit hidden layer with GELU and dropout, and a scalar output layer. Both the fusion block and prediction heads use a dropout rate of 0.1.

We define the valid interval set as $\mathcal{I}_T=\{(i,j):1\leq i\leq j\leq T\}$. A probability distribution over this set is obtained by
\begin{equation}
P(i,j\mid\mathcal{X})=\frac{\exp(s_i+e_j)}{\sum_{(k,l)\in\mathcal{I}_T}\exp(s_k+e_l)},\quad (i,j)\in\mathcal{I}_T.
\label{eq:methods:interval_distribution}
\end{equation}
Invalid intervals and padded positions receive zero probability. Predicted separately for each sequence using both streams, the distribution introduces contiguous-interval structure while allowing uncertainty over interval location and duration. Classification uses soft aggregation weights from the full distribution, without selecting a single interval for hard cropping.

For each volume, the frame-inclusion probability and normalized anatomical aggregation weight are
\begin{equation}
q_t=\sum_{i\leq t\leq j}P(i,j\mid\mathcal{X}),\qquad
\alpha_t=\frac{q_t}{\sum_{u=1}^{T}q_u}.
\label{eq:methods:membership}
\end{equation}
Motion aggregation weights are obtained by averaging neighboring inclusion probabilities and normalizing:
\begin{equation}
r_t=\frac{q_t+q_{t+1}}{2},\qquad
\beta_t=\frac{r_t}{\sum_{u=1}^{T-1}r_u}.
\label{eq:methods:motion_membership}
\end{equation}
The shared interval distribution thus yields weights $\alpha_t$ for anatomical volumes and $\beta_t$ for inter-volume transitions on their respective temporal grids. Small numerical stabilizers are included in the implementation.

\subsubsection{Weighted pooling and evidence fusion}
The temporally weighted representations and full-sequence anatomical descriptor are $z_A=\sum_{t=1}^{T}\alpha_tH_t^A$, $z_M=\sum_{t=1}^{T-1}\beta_tH_t^M$, $z_C=\frac{1}{T}\sum_{t=1}^{T}H_t^A$. The final probability is $p=\sigma\left(g_{\mathrm{cls}}([z_A;z_M;z_C])\right)$.
The classification head $g_{\mathrm{cls}}$ follows the same MLP architecture as the interval prediction heads, but uses 256 hidden units and a dropout rate of 0.2.

The weighted pooling combines with full-sequence anatomical context to provide three complementary descriptors. Because weighting operates on contextualized representations, each descriptor retains access to sequence-wide information; the coefficients specify aggregation weights, not independent diagnostic contributions of raw volumes.

\subsection{Learning Objective and Implementation}
\label{sec:methods:training}
Mitral-AIM is trained end to end using sequence-level binary cross-entropy:
\begin{equation}
\mathcal{L}_{\mathrm{cls}}=-\frac{1}{N}\sum_{n=1}^{N}\left[y_n\log p_n+(1-y_n)\log(1-p_n)\right],
\label{eq:methods:loss}
\end{equation}
where $N$ denotes the number of training sequences and $y_n$ is the diagnostic label inherited from the corresponding patient.

Gradients from the classification objective jointly optimize the anatomical encoder, motion encoder, temporal encoders, and interval-based temporal weighting module. The classification objective supplies the training signal, and the interval distribution constrains how the aggregation weights are formed.

All network components are initialized without external pretraining. Training uses AdamW with a weight decay of $10^{-4}$, five warm-up epochs, and cosine learning-rate decay. The maximum training duration is 120 epochs, with early stopping after 50 epochs without improvement in validation AUROC. Gradient norms are clipped at 1.0. Dropout is set to 0.1 in the temporal encoders and temporal weighting blocks and 0.2 in the classification head. The retained runs use encoder/module learning rates of $10^{-4}/3\times10^{-4}$ and random seeds 42, 43, and 44 for three folds. Data augmentation applies random flips along the two in-plane array axes and a multiplicative intensity gain sampled from $[0.9,1.1]$, followed by clipping to $[0,1]$. The same transformation is applied to every volume within a sequence. Length-aware batching and gradient accumulation accommodate differences in sequence length while retaining complete sequences.

\subsection{Prediction and Evaluation}
\label{sec:methods:evaluation}
Evaluation follows a three-fold patient-grouped protocol with separate training, validation, and test sets. All sequences from the same patient remain within the same partition in each run. For each run, the checkpoint is selected by validation AUROC. Using the selected checkpoint, the classification threshold is determined by maximizing the Youden index on the validation set. The selected checkpoint and threshold are then fixed for test-set evaluation.

Sequence-level AUROC is calculated from predicted probabilities, whereas balanced accuracy (bACC), the unweighted mean of per-class recall, and macro-F1 use the selected fold-specific threshold. Patient-level AUROC uses the mean probability across a patient's sequences. Patient-level classifications are obtained by majority voting over sequence-level decisions; ties are resolved by comparing the mean probability with 0.5 for volumetric models or the corresponding sequence-level threshold for MPR models. Experiments are repeated with three random seeds, each using the same three patient-grouped folds. Metrics are first averaged across folds within each seed, and the mean and sample standard deviation are then calculated across the three seed-level averages.
\input{tables/diagnostic_performance}

\subsection{Comparison Models}
\label{sec:methods:comparators}
Comparators comprise conventional classifiers, volumetric neural models, and MPR-based video models, evaluated using the same patient partitions. Conventional classifiers use sequence-level summary statistics or a shared PyRadiomics representation~\cite{vangriethuysen2017}. Volumetric image models, 3D encoders with sequence aggregation, and MIL comparators share a randomly initialized 3D ResNet-18 encoder~\cite{hara2018}, with aggregation modules adapted to cardiac-cycle features. SwiFT~\cite{kim2023swift} instead models the spatial and temporal axes jointly using its own volumetric Transformer encoder. TDA-Transformer~\cite{ahmadi2024} is adapted without view relevance or Doppler.

SwiFT~\cite{kim2023swift} directly processes the full volume sequence using shifted-window attention over three spatial axes and time. Its volume-sequence adaptation retains every valid $64\times64\times64$ frame, uses $4\times4\times4\times1$ patches and the released architecture's pooled linear head, and is initialized randomly. ASMIL~\cite{ye2026asmil} aggregates the shared 3D ResNet-18 frame features using eight learned feature tokens, 50\% token dropout, and an exponential-moving-average anchor with momentum 0.999.

MPR models use three orthogonal central-plane cine sequences reconstructed from the same volumes. R(2+1)D-18~\cite{tran2018r2plus1d}, Video Swin-T~\cite{liu2022videoswin}, and MViTv2-S~\cite{li2022mvitv2} are trained from random initialization and average view logits before the sigmoid. EchoFM~\cite{kim2025echofm}, EchoPrime~\cite{vukadinovic2026echoprime}, and DISCOVR~\cite{mishra2025discovr} use frozen pretrained encoders with mean view features and linear probes. EchoJEPA~\cite{munim2026echojepa} extends this comparison with a latent predictive echocardiography representation. Its released ViT-L target encoder is frozen; mean view features are standardized using training-set statistics and passed to a linear classifier. Pretrained models keep native preprocessing.

%% file: figures/mitraiam_overview.tex
\begin{figure*}[!t]
\centering
\includegraphics[width=\textwidth]{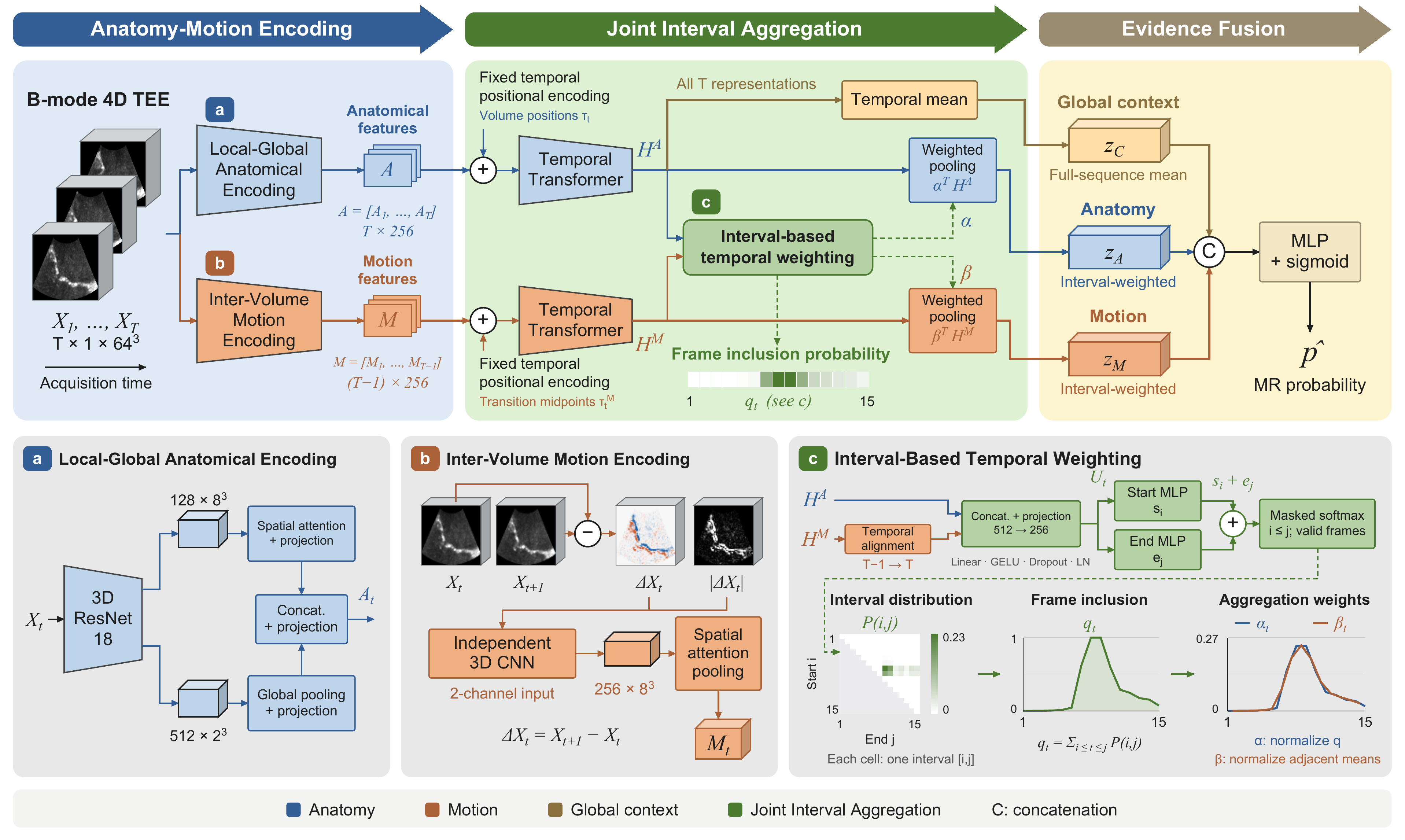}
\caption{Mitral-AIM: overall workflow (top) and module details (bottom).
(a) Local--global anatomical encoding fuses spatially attended intermediate features with deep global features.
(b) Motion encoding processes voxelwise signed and absolute differences between adjacent volumes.
(c) Interval-based temporal weighting combines contextualized anatomy and temporally aligned motion to predict a distribution $P(i,j)$ over valid contiguous intervals $i\leq j$.
Frame-inclusion probabilities $q_t$ and adjacent-frame averages yield normalized anatomical aggregation weights $\alpha_t$ and motion aggregation weights $\beta_t$, respectively.
Weighted pooling combines each feature sequence using its corresponding weights; the resulting descriptors $z_A$ and $z_M$ are concatenated with full-cycle mean anatomical context $z_C$ for MR classification.
$A$ and $M$ stack the per-volume and per-transition tokens $A_t$ and $M_t$ in acquisition order.
Fixed Fourier positional encodings are added at volume positions and transition midpoints before the respective temporal Transformers.
Blue, orange, gold, and green indicate anatomy, motion, global context, and interval aggregation.
$C$ denotes concatenation.}
\label{fig:mitraiam_overview}
\end{figure*}

%% file: tables/dataset_characteristics.tex
\begin{table}[!t]
\caption{Characteristics of the B-mode 4D TEE Dataset}
\label{tab:dataset_characteristics}
\centering
\small
\setlength{\tabcolsep}{3pt}
\renewcommand{\arraystretch}{1.12}
\begin{tabular*}{\columnwidth}{@{\extracolsep{\fill}}lr@{}}
\toprule
Characteristic & Value \\
\midrule
Patient records, $n$ & 264 \\
B-mode 4D TEE sequences, $n$ & 623 \\
Sequence length, volumes & 3--84 \\
Age, years, mean $\pm$ SD & $64.1 \pm 11.7$ \\
Age range, years & 22--87 \\
Male / Female, $n$  & 146 / 118 \\
Nonsignificant / significant MR, sequences, $n$ & 305 / 318 \\
\bottomrule
\end{tabular*}
\par\vspace{3pt}
\end{table}

%% file: tables/diagnostic_performance.tex
\begin{table*}[!t]
\caption{Sequence- and patient-level MR classification from B-mode 4D TEE.}
\label{tab:main}
\centering
\footnotesize
\setlength{\tabcolsep}{3.4pt}
\renewcommand{\arraystretch}{1.16}
\newcommand{\mainmetric}[2]{\mbox{$#1$\,{\fontsize{6}{7}\selectfont$\pm #2$}}}
\newcommand{\cathead}{\multirow{2}{=}{\centering Category}}
\newcommand{\cat}[2]{\multirow{#1}{=}{\raggedright\itshape #2}}
\resizebox{0.9\textwidth}{!}{%
\begin{tabular}{@{}>{\raggedright\arraybackslash}p{2.45cm}lc*{6}{c}@{}}
\toprule
\cathead & \multirow{2}{*}{Method / model} & \multirow{2}{*}{Year} & \multicolumn{2}{c}{AUROC (\%)$\uparrow$} & \multicolumn{2}{c}{bACC (\%)$\uparrow$} & \multicolumn{2}{c}{MF1 (\%)$\uparrow$} \\
\cmidrule(lr){4-5}\cmidrule(lr){6-7}\cmidrule(lr){8-9}
& & & Sequence & Patient & Sequence & Patient & Sequence & Patient \\
\midrule
\cat{4}{Machine learning models} & Radiomics + LR~\cite{vangriethuysen2017} & 2017 & $68.2^{\dagger}$ & $69.6^{\dagger}$ & $65.8^{\dagger}$ & $65.8^{\dagger}$ & $64.9^{\dagger}$ & $65.7^{\dagger}$ \\
& Radiomics + RBF-SVM~\cite{cortes1995} & 1995 & \mainmetric{68.7}{0.0} & \mainmetric{71.7}{0.1} & \mainmetric{67.6}{0.0} & \mainmetric{67.1}{0.3} & \mainmetric{65.6}{0.0} & \mainmetric{65.7}{0.3} \\
& Radiomics + RF~\cite{breiman2001} & 2001 & \mainmetric{65.1}{0.2} & \mainmetric{69.3}{0.1} & \mainmetric{65.0}{0.1} & \mainmetric{67.4}{0.8} & \mainmetric{63.4}{0.3} & \mainmetric{66.4}{1.0} \\
& Radiomics + XGBoost~\cite{chen2016} & 2016 & \mainmetric{62.9}{0.5} & \mainmetric{67.0}{0.7} & \mainmetric{64.5}{1.0} & \mainmetric{64.1}{1.7} & \mainmetric{62.8}{0.7} & \mainmetric{62.9}{1.4} \\
\midrule
\cat{3}{Image-based models} & Mean-volume 3D ResNet-18~\cite{hara2018} & 2018 & \mainmetric{67.2}{0.9} & \mainmetric{69.7}{2.1} & \mainmetric{65.2}{1.4} & \mainmetric{67.6}{0.6} & \mainmetric{64.4}{1.3} & \mainmetric{65.7}{1.5} \\
& Frame-probability 3D ResNet-18~\cite{hara2018} & 2018 & \mainmetric{69.6}{0.7} & \mainmetric{70.5}{2.1} & \mainmetric{66.8}{1.1} & \mainmetric{65.7}{2.9} & \mainmetric{66.7}{1.3} & \mainmetric{65.1}{2.8} \\
& Feature-mean 3D ResNet-18~\cite{hara2018} & 2018 & \mainmetric{66.0}{1.4} & \mainmetric{67.7}{2.1} & \mainmetric{64.6}{0.8} & \mainmetric{65.4}{0.3} & \mainmetric{63.3}{2.0} & \mainmetric{64.6}{0.8} \\
\midrule
\cat{5}{3D encoders with sequence aggregation} & 3D ResNet--attention pooling~\cite{ilse2018} & 2018 & \mainmetric{66.1}{2.9} & \mainmetric{67.7}{2.5} & \mainmetric{65.4}{2.9} & \mainmetric{66.5}{2.9} & \mainmetric{64.7}{2.8} & \mainmetric{64.1}{3.3} \\
& 3D ResNet--BiLSTM~\cite{graves2005} & 2005 & \mainmetric{68.2}{1.8} & \mainmetric{69.8}{2.3} & \mainmetric{65.7}{1.3} & \mainmetric{65.3}{1.9} & \mainmetric{64.6}{2.0} & \mainmetric{64.5}{2.1} \\
& 3D ResNet--BiGRU~\cite{cho2014} & 2014 & \mainmetric{66.4}{0.6} & \mainmetric{68.6}{0.9} & \mainmetric{64.9}{0.2} & \mainmetric{65.7}{1.1} & \mainmetric{63.6}{0.6} & \mainmetric{63.9}{0.3} \\
& 3D ResNet--Transformer~\cite{vaswani2017} & 2017 & \mainmetric{62.1}{0.8} & \mainmetric{61.6}{3.2} & \mainmetric{61.4}{1.0} & \mainmetric{63.7}{2.1} & \mainmetric{59.9}{0.2} & \mainmetric{60.9}{0.8} \\
& 3D ResNet--dilated TCN~\cite{bai2018tcn} & 2018 & \mainmetric{67.9}{1.7} & \mainmetric{68.0}{1.5} & \mainmetric{65.7}{1.2} & \mainmetric{64.5}{1.4} & \mainmetric{64.8}{1.2} & \mainmetric{63.8}{1.1} \\
\midrule
\cat{2}{3D spatiotemporal models} & SwiFT~\cite{kim2023swift} & 2023 & \mainmetric{66.8}{2.0} & \mainmetric{68.6}{1.8} & \mainmetric{64.5}{1.9} & \mainmetric{65.3}{1.8} & \mainmetric{63.4}{2.2} & \mainmetric{63.2}{2.1} \\
& TDA-Transformer~\cite{ahmadi2024} & 2024 & \mainmetric{63.2}{1.7} & \mainmetric{64.0}{0.6} & \mainmetric{62.3}{1.5} & \mainmetric{63.0}{0.9} & \mainmetric{60.8}{1.0} & \mainmetric{61.1}{1.3} \\
\midrule
\cat{9}{Weakly supervised MIL aggregators} & DSMIL~\cite{li2021dsmil} & 2021 & \mainmetric{68.7}{2.0} & \mainmetric{69.8}{2.6} & \mainmetric{66.5}{2.0} & \mainmetric{67.3}{2.2} & \mainmetric{64.8}{2.5} & \mainmetric{65.2}{3.0} \\
& TransMIL temporal adaptation~\cite{shao2021transmil} & 2021 & \mainmetric{67.0}{1.6} & \mainmetric{67.6}{1.7} & \mainmetric{65.0}{1.3} & \mainmetric{65.6}{1.5} & \mainmetric{63.6}{1.3} & \mainmetric{62.4}{1.9} \\
& CLAM-SB~\cite{lu2021clam} & 2021 & \mainmetric{67.9}{1.6} & \mainmetric{68.1}{4.8} & \mainmetric{66.3}{1.8} & \mainmetric{66.4}{3.7} & \mainmetric{65.0}{2.6} & \mainmetric{64.5}{4.5} \\
& DTFD-MIL~\cite{zhang2022dtfdmil} & 2022 & \mainmetric{69.2}{1.4} & \mainmetric{69.4}{1.9} & \mainmetric{65.9}{1.0} & \mainmetric{65.6}{0.8} & \mainmetric{64.3}{1.5} & \mainmetric{63.8}{1.0} \\
& ILRA-MIL aggregator~\cite{xiang2023ilra} & 2023 & \mainmetric{69.1}{1.9} & \mainmetric{70.1}{2.7} & \mainmetric{67.1}{1.5} & \mainmetric{67.7}{1.4} & \mainmetric{66.3}{1.7} & \mainmetric{66.6}{1.6} \\
& ACMIL-GA~\cite{zhang2024acmil} & 2024 & \mainmetric{69.2}{1.9} & \mainmetric{71.1}{1.1} & \mainmetric{66.2}{2.6} & \mainmetric{66.8}{2.4} & \mainmetric{64.7}{3.8} & \mainmetric{65.3}{4.7} \\
& ACMIL-MHA~\cite{zhang2024acmil} & 2024 & \mainmetric{66.5}{2.8} & \mainmetric{66.5}{2.9} & \mainmetric{64.8}{1.6} & \mainmetric{64.9}{2.5} & \mainmetric{63.5}{2.7} & \mainmetric{63.0}{3.6} \\
& WiKG temporal-graph adaptation~\cite{li2024wikg} & 2024 & \mainmetric{68.9}{1.3} & \mainmetric{70.5}{1.2} & \mainmetric{65.8}{1.2} & \mainmetric{64.9}{1.9} & \mainmetric{64.5}{1.9} & \mainmetric{64.2}{2.5} \\
& ASMIL (3D feature aggregation)~\cite{ye2026asmil} & 2026 & \mainmetric{68.3}{0.7} & \mainmetric{70.2}{1.5} & \mainmetric{66.2}{1.3} & \mainmetric{66.2}{1.8} & \mainmetric{65.1}{1.1} & \mainmetric{65.6}{2.2} \\
\midrule
\cat{3}{Video backbones (MPR input)} & R(2+1)D-18~\cite{tran2018r2plus1d} & 2018 & \mainmetric{\ranksecond{71.7}}{3.1} & \mainmetric{\ranksecond{72.3}}{4.8} & \mainmetric{67.6}{2.7} & \mainmetric{68.3}{3.0} & \mainmetric{66.7}{3.8} & \mainmetric{65.1}{4.8} \\
& Video Swin-T~\cite{liu2022videoswin} & 2022 & \mainmetric{60.5}{0.5} & \mainmetric{60.5}{0.7} & \mainmetric{61.6}{0.5} & \mainmetric{60.9}{0.4} & \mainmetric{60.3}{1.1} & \mainmetric{59.1}{1.8} \\
& MViTv2-S~\cite{li2022mvitv2} & 2022 & \mainmetric{58.4}{1.2} & \mainmetric{61.0}{0.8} & \mainmetric{59.0}{0.6} & \mainmetric{59.6}{0.5} & \mainmetric{56.8}{0.6} & \mainmetric{57.5}{0.8} \\
\midrule
\cat{4}{Echocardiography-pretrained encoders (MPR input)} & EchoFM (ViT-L + linear probe)~\cite{kim2025echofm} & 2025 & \mainmetric{67.2}{0.7} & \mainmetric{69.6}{0.6} & \mainmetric{65.4}{1.1} & \mainmetric{66.3}{1.2} & \mainmetric{63.2}{1.3} & \mainmetric{64.0}{0.2} \\
& EchoPrime (MViTv2-S + linear probe)~\cite{vukadinovic2026echoprime} & 2026 & \mainmetric{70.2}{0.7} & \mainmetric{71.0}{0.7} & \mainmetric{67.0}{0.9} & \mainmetric{\ranksecond{69.0}}{0.4} & \mainmetric{65.7}{1.0} & \mainmetric{\ranksecond{68.9}}{0.3} \\
& DISCOVR (VideoMAE-B + linear probe)~\cite{mishra2025discovr} & 2025 & \mainmetric{61.7}{0.2} & \mainmetric{64.7}{0.0} & \mainmetric{61.0}{0.4} & \mainmetric{61.8}{1.8} & \mainmetric{58.7}{1.5} & \mainmetric{58.8}{1.9} \\
& EchoJEPA (ViT-L + linear probe)~\cite{munim2026echojepa} & 2026 & \mainmetric{71.2}{0.8} & \mainmetric{70.9}{1.0} & \mainmetric{\ranksecond{68.1}}{0.3} & \mainmetric{67.7}{0.8} & \mainmetric{\ranksecond{67.2}}{0.1} & \mainmetric{67.5}{0.5} \\
\midrule
\rowcolor{black!9}
\itshape Proposed & \textbf{Mitral-AIM (Ours)} & -- & \mainmetric{\rankfirst{72.7}}{0.5} & \mainmetric{\rankfirst{76.7}}{1.9} & \mainmetric{\rankfirst{70.0}}{1.8} & \mainmetric{\rankfirst{72.3}}{1.5} & \mainmetric{\rankfirst{69.8}}{1.8} & \mainmetric{\rankfirst{71.2}}{2.4} \\
\bottomrule
\end{tabular}}
\par\smallskip
\parbox{\textwidth}{\footnotesize $^{\dagger}$Three-fold mean; seed SD is not applicable. \textcolor{TableBest}{\textbf{Bold red}} and \textcolor{TableSecond}{\underline{blue underlined}} means indicate the highest and second-highest values in each column, respectively.}
\end{table*}

%% file: sections/results.tex
\section{Results}
\label{sec:results}
\subsection{Diagnostic Performance}
\label{sec:results:diagnostic}

Mitral-AIM achieved the highest mean values for all metrics reported in Table~\ref{tab:main}. Patient-level AUROC was $76.7\pm1.9\%$, exceeding R(2+1)D-18 ($72.3\pm4.8\%$), the strongest comparator on this metric, by 4.4 percentage points. Sequence-level AUROC was $72.7\pm0.5\%$, 1.0 percentage point above the same model ($71.7\pm3.1\%$). Patient-level bACC and MF1 were $72.3\pm1.5\%$ and $71.2\pm2.4\%$, respectively, exceeding the next-highest means, both achieved by EchoPrime, by 3.3 and 2.3 percentage points. The higher mean performance therefore extended from diagnostic ranking to threshold-dependent classification.

Among volumetric comparators, Radiomics + RBF-SVM achieved the highest patient-level AUROC ($71.7\pm0.1\%$), while ACMIL-GA led the neural models ($71.1\pm1.1\%$). \mbox{Mitral-AIM} exceeded these means by 5.0 and 5.6 percentage points, respectively. Frame-probability 3D ResNet-18 achieved the highest sequence-level AUROC in this group ($69.6\pm0.7\%$), 3.1 percentage points below Mitral-AIM. SwiFT and ASMIL reached patient-level AUROCs of $68.6\pm1.8\%$ and $70.2\pm1.5\%$, 8.1 and 6.5 points below Mitral-AIM.

Among pretrained encoders evaluated with linear probes, EchoJEPA achieved the highest sequence-level AUROC ($71.2\pm0.8\%$), with a patient-level AUROC of $70.9\pm1.0\%$. EchoPrime achieved the highest patient-level mean in this group ($71.0\pm0.7\%$). Mitral-AIM exceeded EchoJEPA by 1.5 and 5.8 percentage points at the sequence and patient levels, respectively, extending the comparison to representations learned through latent predictive echocardiography pretraining.

\subsection{Temporal Concentration and Functional Relevance}
\label{sec:results:interpretation}
\label{sec:results:temporal}

\input{tables/temporal_localization}

Expert systole was annotated on 20 test-set sequences from 20 patients to test whether learned aggregation weights concentrate on this phase. Table~\ref{tab:temporal_localization} compares three quantities on this subset. Systolic mass is the sum of normalized aggregation weights within expert systole. Systolic enrichment divides this mass by the fraction of frames in systole. A value of one means the weight placed in systole equals the share of frames it occupies; larger values indicate concentration beyond that share. Peak in systole records whether the maximum-weight frame lies within expert systole.

Mitral-AIM assigned a mean systolic mass of $0.868\pm0.178$ and a mean enrichment of $1.755\pm0.538$. This enrichment means the weight placed in expert systole was $1.755$ times the fraction of frames that phase occupied. Mean comparator masses ranged from $0.401$ to $0.568$, with enrichment near or below one. Paired systolic-mass tests favored Mitral-AIM over all five comparators (two-sided Wilcoxon tests, all $p<0.001$).

The maximum-weight frame fell within expert systole in 19 of 20 sequences for Mitral-AIM and in 6 to 12 of 20 sequences for the comparators (exact paired McNemar tests, all $p\leq0.039$). Fig.~\ref{fig:temporal_cases} shows one sequence from each MR class. In both, the highest Mitral-AIM weights fall inside expert systole, on frames 2--3 in the nonsignificant case and frames 3--6 in the significant case.

\input{figures/representative_temporal_cases}

\input{tables/functional_deletion}

Table~\ref{tab:temporal_deletion} tests whether the learned summary interval is useful for classification. Summary-interval deletion removes that interval. Matched random deletion removes an equally long interval elsewhere and averages 50 such deletions for each sequence, separating the interval's content from the effect of deleting the same number of frames. Sequence-level AUROC fell from $72.7\pm0.5\%$ to $63.2\pm2.0\%$ after summary-interval deletion, $4.3$ percentage points below matched random deletion ($67.5\pm3.0\%$), as did bACC and MF1.

\subsection{Local--Global Anatomical Representation Analysis}
\label{sec:results:spatial}

\input{figures/spatial_representation_similarity}

The frame-wise similarity matrices in Fig.~\ref{fig:spatial_representations} illustrate how local and global anatomical features describe variation within a 4D TEE sequence. Pairwise cosine similarity was computed from the projected features before temporal position encoding and Transformer processing in two representative sequences.

In Example A, global appearance is largely preserved: off-diagonal cosine similarities remain $0.818$--$0.999$. Within that stable context, local features separate frames 1--3 from frames 4--6, and off-diagonal local similarities fall to $0.229$. The local branch retains phase-dependent anatomical distinctions that remain unresolved by global pooling. In Example B, both representations assign similar features to the frames inside expert systole (frames 5--8) and keep that group distinct from frames 1--4. Frame 10 rejoins them, so systole is coherent.

The interval weights follow this organization. In Example A, the weight rises inside expert systole (frames 2--4) and peaks at frame 3 ($0.305$). In Example B, frames 1--4, whose appearance differs from systole, receive almost no weight, whereas the weight increases across the systolic frames. Aggregation is therefore concentrated on the coherent systolic representation.

\subsection{Ablation Study}
\label{sec:results:analysis}

\input{tables/ablation}

Table~\ref{tab:ablation} removes one component of Mitral-AIM at a time. Anatomical variants omit the local branch, the global branch, or both; the motion variant omits inter-volume encoding; and the interval variant keeps temporal encoding, replacing learned interval weights with uniform aggregation. Relative to the full model ($72.7\pm0.5\%$ sequence-level AUROC and $76.7\pm1.9\%$ patient-level AUROC), these removals gave $69.5\pm4.1\%$ and $68.8\pm3.7\%$ without the local branch, $69.5\pm0.5\%$ and $70.6\pm3.7\%$ without the global branch, and $67.8\pm2.2\%$ and $69.7\pm4.5\%$ without both. Uniform aggregation gave $69.7\pm1.0\%$ and $71.9\pm4.1\%$. Removing the motion stream produced the largest drop, to $66.5\pm2.3\%$ and $68.2\pm1.2\%$.

%% file: tables/temporal_localization.tex
\begin{table}[!htb]
\caption{Temporal evidence within expert-defined systole on 20 expert-annotated test-set sequences.}
\label{tab:temporal_localization}
\centering
\footnotesize
\setlength{\tabcolsep}{3.2pt}
\renewcommand{\arraystretch}{1.12}
\newcommand{\locmetric}[2]{\mbox{$#1$\,{\fontsize{6}{7}\selectfont$\pm #2$}}}
\resizebox{\columnwidth}{!}{%
\begin{tabular}{@{}lccc@{}}
\toprule
Method & Systolic mass $\uparrow$ & Systolic enrichment $\uparrow$ & Peak in systole $\uparrow$ \\
\midrule
ACMIL-GA~\cite{zhang2024acmil} & \locmetric{0.401}{0.247} & \locmetric{0.759}{0.402} & 6/20 \\
3D ResNet--attention pooling~\cite{ilse2018} & \locmetric{0.517}{0.158} & \locmetric{0.989}{0.078} & 10/20 \\
WiKG temporal-graph adaptation~\cite{li2024wikg} & \locmetric{0.525}{0.151} & \locmetric{1.006}{0.023} & 12/20 \\
TDA-Transformer~\cite{ahmadi2024} & \locmetric{0.561}{0.340} & \locmetric{1.036}{0.580} & 10/20 \\
CLAM-SB~\cite{lu2021clam} & \locmetric{0.568}{0.168} & \locmetric{1.110}{0.250} & 12/20 \\
\rowcolor{black!9}
\textbf{Mitral-AIM (Ours)} & \locmetric{\mathbf{0.868}}{0.178} & \locmetric{\mathbf{1.755}}{0.538} & \textbf{19/20} \\
\bottomrule
\end{tabular}}
\end{table}

%% file: figures/representative_temporal_cases.tex
\begin{figure*}[!t]
\centering
\includegraphics[width=\textwidth]{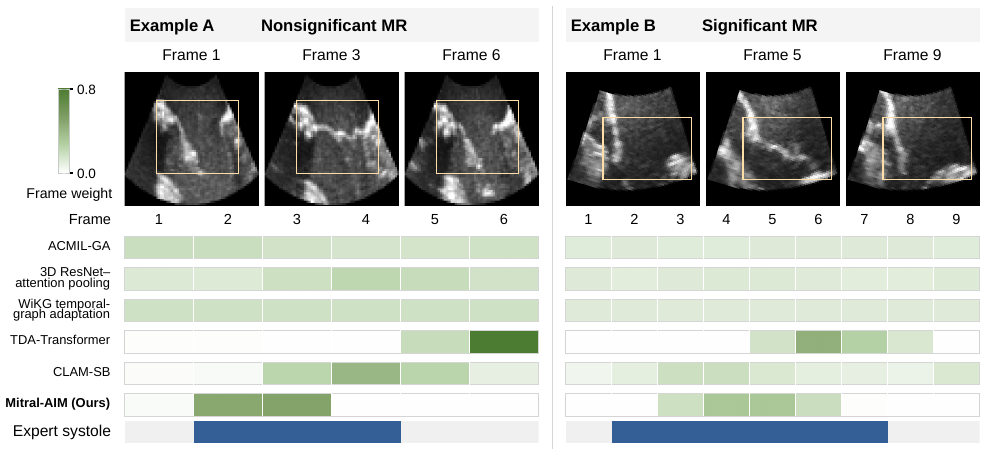}
\caption{Temporal evidence in representative (A) nonsignificant and (B) significant MR cases from the annotated test sequences.
Each case displays three reconstructed slices from a fixed plane.
Heatmaps show all valid frames for the methods in Table~\ref{tab:temporal_localization}, in the same order, with weights normalized to sum to one per method and sequence and a shared linear color scale.
Blue bars mark expert systole.
Yellow rectangles are manually placed visual guides.}
\label{fig:temporal_cases}
\end{figure*}

%% file: tables/functional_deletion.tex
\begin{table}[!htb]
\caption{Effect of deleting model-derived summary intervals on sequence-level MR classification.}
\label{tab:temporal_deletion}
\centering
\footnotesize
\setlength{\tabcolsep}{3.2pt}
\renewcommand{\arraystretch}{1.12}
\newcommand{\delmetric}[2]{\mbox{$#1$\,{\fontsize{6}{7}\selectfont$\pm #2$}}}
\resizebox{0.85\columnwidth}{!}{%
\begin{tabular}{@{}lccc@{}}
\toprule
Condition & AUROC (\%)$\uparrow$ & bACC (\%)$\uparrow$ & MF1 (\%)$\uparrow$ \\
\midrule
Summary interval deletion & \delmetric{63.2}{2.0} & \delmetric{53.3}{1.2} & \delmetric{43.2}{4.9} \\
Matched random deletion & \delmetric{67.5}{3.0} & \delmetric{56.1}{3.4} & \delmetric{47.9}{6.9} \\
\rowcolor{black!9}
\textbf{Original} & \delmetric{\mathbf{72.7}}{0.5} & \delmetric{\mathbf{70.0}}{1.8} & \delmetric{\mathbf{69.8}}{1.8} \\
\bottomrule
\end{tabular}}
\end{table}

%% file: figures/spatial_representation_similarity.tex
\begin{figure*}[!t]
\centering
\includegraphics[width=0.85\textwidth]{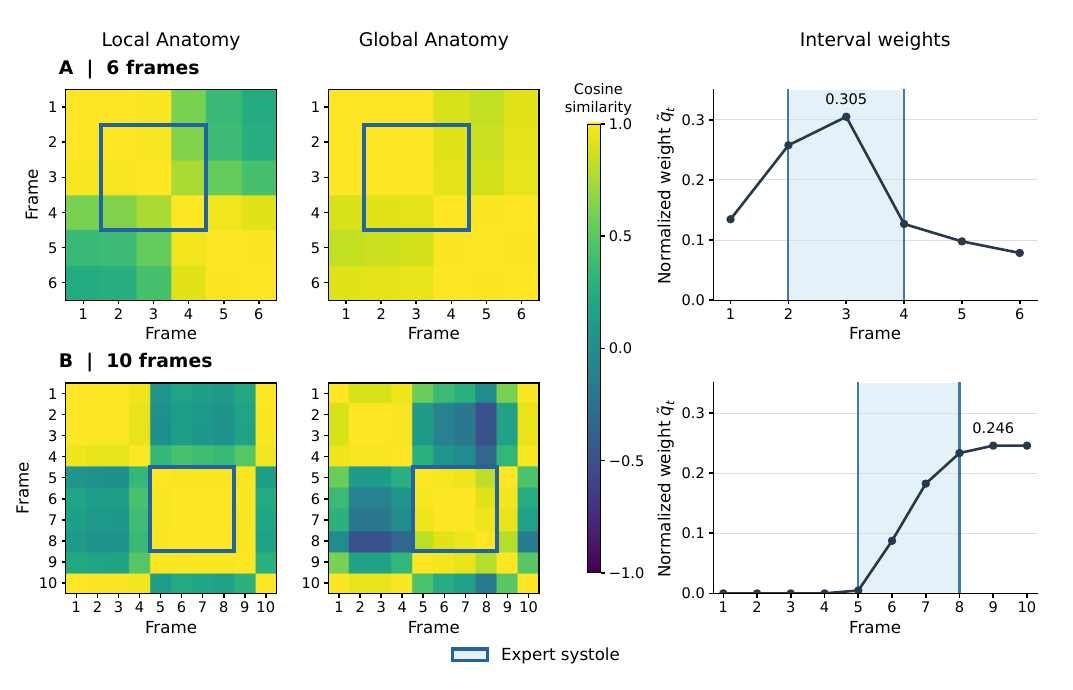}
\caption{Local--global anatomical similarity and interval weights in two representative sequences: (A) six frames and (B) ten frames.
The first two columns show pairwise cosine similarity of projected local and global features before temporal encoding, using a shared $[-1,1]$ color scale.
The right column shows normalized interval weights $\tilde q_t=q_t/\sum_u q_u=\alpha_t$ from the joint anatomy--motion interval distribution.
Blue outlines in the matrices mark expert systole. In the right column, two blue vertical lines mark its start and end frames, with light-blue shading between them: frames 2--4 in A and frames 5--8 in B.}
\label{fig:spatial_representations}
\end{figure*}

%% file: tables/ablation.tex
\begin{table}[!htb]
\caption{Ablation of anatomical, motion, and temporal components.}
\label{tab:ablation}
\centering
\footnotesize
\setlength{\tabcolsep}{4pt}
\renewcommand{\arraystretch}{1.12}
\newcommand{\ablmetric}[2]{\mbox{$#1$\,{\fontsize{6}{7}\selectfont$\pm #2$}}}
\resizebox{0.95\columnwidth}{!}{%
\begin{tabular}{@{}lcc@{}}
\toprule
Configuration & Sequence AUROC (\%)$\uparrow$ & Patient AUROC (\%)$\uparrow$ \\
\midrule
\rowcolor{black!9}
\textbf{Mitral-AIM (full)} & \ablmetric{\mathbf{72.7}}{0.5} & \ablmetric{\mathbf{76.7}}{1.9} \\
w/o local anatomical branch & \ablmetric{69.5}{4.1} & \ablmetric{68.8}{3.7} \\
w/o global anatomical branch & \ablmetric{69.5}{0.5} & \ablmetric{70.6}{3.7} \\
w/o local and global anatomy & \ablmetric{67.8}{2.2} & \ablmetric{69.7}{4.5} \\
w/o motion stream & \ablmetric{66.5}{2.3} & \ablmetric{68.2}{1.2} \\
w/o interval weighting module & \ablmetric{69.7}{1.0} & \ablmetric{71.9}{4.1} \\
\bottomrule
\end{tabular}}
\end{table}

%% file: sections/discussion.tex
\section{Discussion}
\label{sec:discussion}

\textbf{Task-specific modeling for volumetric MR detection.}
Direct MR detection from B-mode 4D TEE benefits from modeling how diagnostic information is distributed in a volumetric sequence. Spatiotemporal models, MIL aggregators, and frozen echocardiographic representations trailed Mitral-AIM at the patient level, motivating localized anatomical detail, inter-volume change through feature extraction, and aggregation of observations whose relevance varies over time. Removing explicit inter-volume encoding reduced sequence- and patient-level AUROC by 6.2 and 8.5 percentage points (Table~\ref{tab:ablation}), so dynamic information contributed beyond the anatomical pathway. Encoders, input representations, and optimization also differ, so these comparisons do not isolate design choices. Frozen MPR linear probes evaluate transfer of pretrained representations in this setting, not their potential after full adaptation to volumetric TEE.

\textbf{Systolic alignment and predictive relevance.}
Without phase annotations, patient-level supervision aligned temporal weights with expert-defined systole. Mitral-AIM placed its maximum weight within systole in 19 of 20 annotated sequences, compared with 6 to 12 for the temporal comparators (Table~\ref{tab:temporal_localization}). Systolic enrichment shows that this concentration exceeded the fraction of frames occupied by systole, so learning from B-mode sequences can emphasize observations relevant to systolic valve closure. In both MR classes, the model may use a common phase in which anatomical and dynamic differences are informative, rather than treating systole as a disease marker. Summary-interval deletion reduced AUROC by 9.5 percentage points, compared with 5.2 after equally long random deletion (Table~\ref{tab:temporal_deletion}), supporting temporal summaries as cues for reviewing grayscale acquisitions. Weights on contextualized features, and deletion that perturbs the input, show temporal alignment and prediction sensitivity without a causal account of regurgitation or a validated boundary of pathological activity.

\textbf{Complementary anatomical representations.}
Local features preserved distinctions unresolved by the global representation. With relatively stable global appearance, the local branch separated portions of the cardiac cycle (Fig.~\ref{fig:spatial_representations}), consistent with intermediate-resolution detail when relevant structures occupy little of the volume. Recurring configurations in both branches indicate appearance similarity beyond temporal adjacency. Removing either anatomical branch lowered sequence-level AUROC by 3.2 percentage points, and removing both by 4.9 (Table~\ref{tab:ablation}).

\textbf{Limitations and future validation.}
The dataset contains 623 sequences from 264 patients in one cohort on one ultrasound platform. Repeated acquisitions add observations, but independent patients remain limited, and internal evaluation cannot establish robustness across institutions, equipment, acquisition practice, or populations. External multicenter and cross-device validation is the principal next step, followed by subgroup analysis of severity, etiology, and image quality where labels are available. Temporal validation is limited to 20 patients and shows nonuniform systolic enrichment, so larger annotated samples are needed to characterize when the temporal and spatial representations remain informative and when they fail. Separating nonsignificant from clinically significant MR does not establish full severity grading. B-mode predictions reflect anatomical and dynamic correlates of the label without quantifying regurgitant flow, and may complement comprehensive echocardiographic assessment~\cite{zoghbi2017}. Prospective reader studies are needed to determine whether predictions and temporal summaries improve clinical review.

%% file: sections/conclusion.tex
\section{Conclusion}
\label{sec:conclusion}
In this paper, we propose Mitral-AIM, an interval-aware anatomy--motion framework for assessing MR severity from B-mode 4D TEE. To our knowledge, it is the first model to assess MR severity on four-dimensional echocardiographic data, and it does not require color Doppler input. Local--global anatomical encoding and independent inter-volume motion encoding provide complementary representations of each sequence, while interval-based temporal weighting derives coordinated aggregation weights for both streams from a distribution over contiguous intervals. Experiments demonstrate that Mitral-AIM outperforms the compared methods, showcasing its effectiveness for direct MR assessment from grayscale volumetric sequences.